\documentclass[letterpaper, 10 pt, conference]{ieeeconf}
\IEEEoverridecommandlockouts
\usepackage{epsfig}
\usepackage{times}
\usepackage{amsmath}
\usepackage{amssymb}
\usepackage{multirow}
\usepackage[T1]{fontenc}  
\usepackage[utf8]{inputenc}
\usepackage[english]{babel}
\usepackage[export]{adjustbox}
\usepackage{graphicx}
\usepackage{blindtext}
\usepackage{booktabs, threeparttable}
\usepackage{subcaption}
\usepackage[colorinlistoftodos, german]{todonotes}
\usepackage[T1]{fontenc}
\usepackage[utf8]{inputenc}
\usepackage{textcomp}
\usepackage[absolute,showboxes]{textpos}
\usepackage{hyperref}

\newcommand{\vx}{\mathbf{x}}
\newcommand{\abs}[1]{\left\lvert#1\right\rvert}
\newcommand{\npin}{N_\mathrm{pin}}
\newcommand{\svis}{s_\mathrm{vis}}
\newcommand{\smin}{s_\mathrm{min}}
\newcommand{\svisproj}{s_{\mathrm{vis}}^{(e \rightarrow o)}}
\newcommand{\Sfull}{S_\mathrm{full} }
\newcommand{\sfull}{s_\mathrm{full} }
\newcommand{\aacc}{a_\mathrm{acc} }
\newcommand{\adec}{a_\mathrm{dec} }
\newcommand{\acft}{a_\mathrm{cft} }

\newcommand{\vdes}{v_\mathrm{des} }
\newcommand{\vrel}{v_\mathrm{rel} }

\newcommand{\xMP}{x_\mathrm{MP}}

\newcommand{\aidm}{a_\mathrm{IDM} }

\newcommand{\tsafe}{t_\mathrm{safe} }
\newcommand{\mergepoint}{$\mathrm{MP}$}
\hypersetup{
    colorlinks=true,
    linkcolor=black,
    citecolor=black,
    filecolor=black,
    urlcolor=black,
pdfinfo={
Title={Limited Visibility and Uncertainty Aware Motion Planning for Automated Driving},
Author={Omer Sahin Tas},
Subject={IEEE Intelligent Vehicle Symposium 2018}
}
}
\TPMargin{5pt}
\newcommand{\copyrightstatement}{
    \begin{textblock}{0.84}(0.08,0.02)
         \noindent{\footnotesize{\copyright  2018 IEEE.
         Personal use of this material is permitted. Permission from IEEE must be obtained for all other uses, in any current or future media, including reprinting/republishing this material for advertising or promotional purposes, creating new collective works, for resale or redistribution to servers or lists, or reuse of any copyrighted component of this work in other works. DOI: \url{https://doi.org/10.1109/IVS.2018.8500369}
         
         \vspace{2mm}
         \noindent
         In Proceedings of the IEEE Intelligent Vehicles Symposium (IV), Changshu-Suzhou China, 26-30 June 2018, pp.\ 1171--1178}}
    \end{textblock}
}

\title{\LARGE \bf 
Limited Visibility and Uncertainty Aware Motion Planning for Automated Driving}

\author{
Ömer Şahin Taş 
and Christoph Stiller
\thanks{
The authors are with 
FZI Research Center for Information Technology
at the Karlsruhe Institute of Technology, 
Haid-und-Neu-Str. 10-14, 76131 Karlsruhe, Germany.
E-mail: 
{\tt\small 
\{\texttt{tas}, 
\texttt{stiller\}@fzi.de}
}
}
}

\begin{document}
\copyrightstatement
\maketitle
\thispagestyle{empty}
\pagestyle{empty}

\begin{abstract} 
Adverse weather conditions and occlusions in urban environments result in impaired perception. 
The uncertainties are handled in different modules of an automated vehicle, 
ranging from sensor level over situation prediction until motion planning.
This paper focuses on motion planning given an uncertain environment model with occlusions. 
We present a method to remain collision free for the worst-case evolution of the given scene.
We define criteria that measure the available margins to a collision while considering visibility and interactions
and consequently integrate conditions that apply these criteria into an optimization-based motion planner.
We show the generality of our method by validating it in several distinct urban scenarios.
\end{abstract}

\section{Introduction}
\label{sec:introduction}

Automated driving in urban environments and under harsh conditions is both challenging 
and remains on top of research. 
Adverse weather and light conditions result in poor quality in perception 
by introducing higher uncertainties and limited receptive field. 
Urban environments are further bound by occlusions and are subject to unexpected scene evolutions 
making a reliable situation prediction more difficult. 
The operability of an automated vehicle under these conditions can be maintained by a well-developed perception system 
utilizing different measurement principles and benefits from redundancy. 
However, even state-of-the art automated vehicles equipped with diverse and advanced sensor setups 
cannot deal with the aforementioned problems alone and yield imperfect results. 
It is the duty of the planning modules to consider the remaining uncertainties
and plan motion that compensate those deficiencies.  
 
The uncertainties that accumulate up to motion planning comprise the ones 
that originate from sensor measurements, scene understanding and prediction modules. 
Even with a heterogenous sensor setup, the fusion module can 
yield object data with high variance and also miss objects in the current scene due to occlusions and insufficient resolutions in the distance. 
The offline map, which serves as a basis for scene understanding can be outdated. 
The future prediction of the current scene may misclassify maneuver intentions
and yield unreliable future predictions. 

\begin{figure}
\vspace{2mm}
\includegraphics[width=\columnwidth]{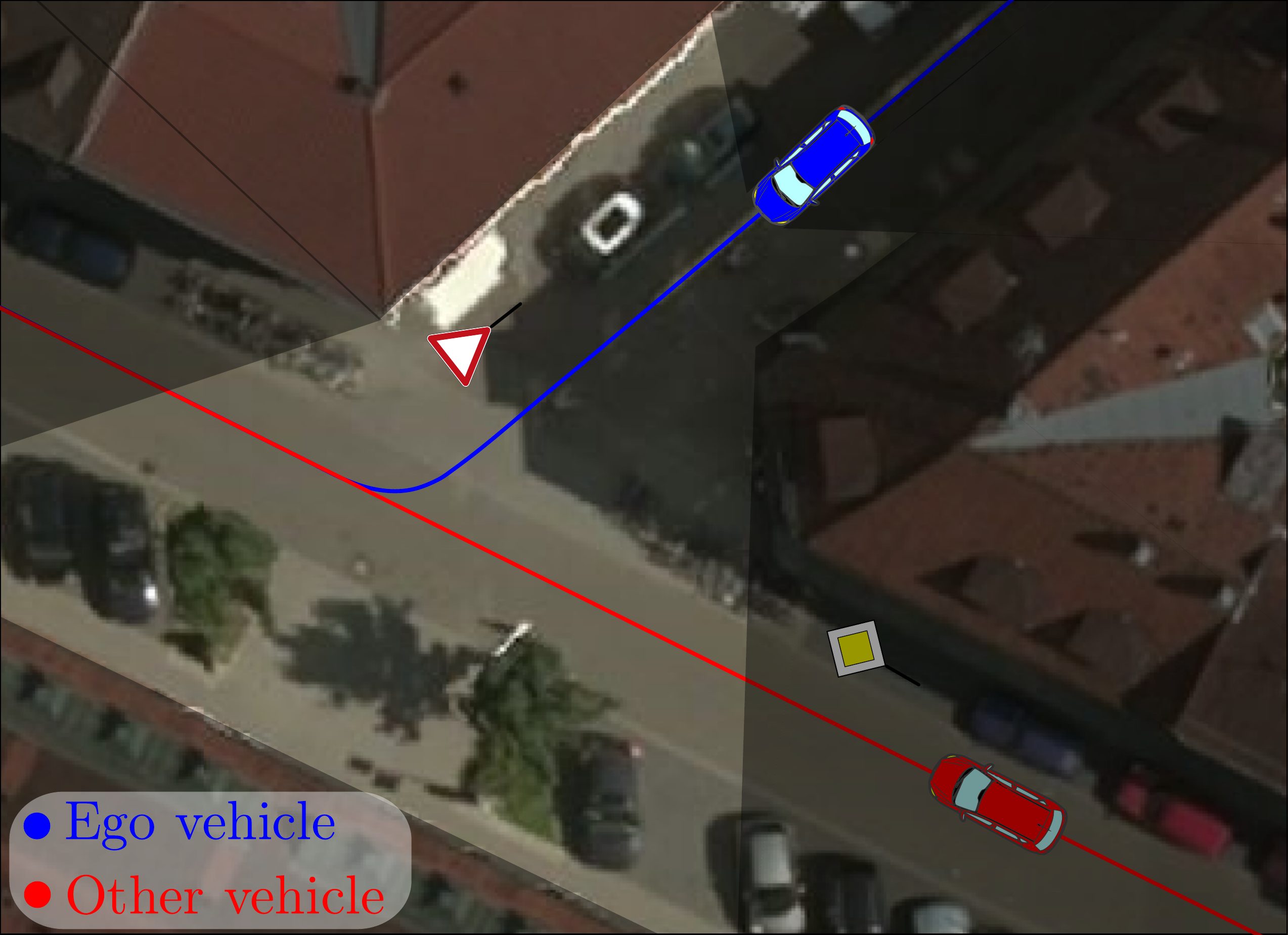}
\caption{An exemplary intersection scenario highlighting a use case of the proposed approach. 
The ego vehicle, depicted in blue throughout this paper, proceeds to an intersection at which it has to yield to. 
The vehicle has limited visibility and an intersecting vehicle is approaching from the occluded region.
The planner of the vehicle has to consider that there might be a vehicle approaching, 
and adequately reduce its speed.
}
\label{fig:figure_01}
\end{figure}

In this paper we aim to bridge the gap between a incomplete environment model and motion planning.
We introduce methodology to remain collision free while considering
uncertain perception, limited visibility and possible uncompliant behavior. 
As we focus on motion planning, we make simplifying assumptions on sensor range and available offline map data. 
We observe the probable maneuver alternatives of other participants by modeling interactions and 
by incorporating worst-case hypothesis, and in this way advance from reactive to proactive planning,
\textit{cf.\ }Fig.\ \ref{fig:figure_01}. 
We demonstrate the success of our approach in closed-loop simulation.

The rest of the paper is structured as follows: 
In Section \ref{sec:related_work} we give an overview of the related work.
Subsequently, in Section \ref{sec:environment_model}, 
we present the environment model together with the simplifying assumptions to which the planning is done for. 
Once the environment model is defined, we continue with Section \ref{sec:robust_planning}, 
in which we briefly present the utilized motion planner and then define 
\textit{safe motion planning} from the perspective of automated driving.
We subsequently introduce conditions to consider these in a motion planner. 
Although we derive these constraints for an optimization based planner, 
these are applicable to any planner that is replans its motion. 
We continue with demonstrating the success of the proposed approach in Section \ref{sec:experiments}.
Section \ref{sec:conclusion} concludes the paper by summarizing out our contribution and providing future research directives.

\section{Related Work}
\label{sec:related_work}

Safe motion planning requires to consider accumulating uncertainties.
Many works adressed different aspects of criticality assessment and safe planning. 

An overview on time metrics (TTX) is presented in \cite{hillenbrand2006efficient}. 
These time metrics are however not robust to uncertain information and do not incorporate interactions of other traffic participants. 
In order to increase robustness and efficiency, 
\cite{berthelot2012novel}
uses stochastic linearization.
They utilize unscented transformation 
to model Gaussian uncertainties in perception and prediction,
and calculate a collision probability.
An in depth analysis for the propagation of estimation and prediction uncertainties 
and the therefrom resulting criticality measures is given in \cite{stellet2015uncertainty}. 
The work derives closed form expressions and in a later work,
the authors determine therefrom resulting capabilities 
of an emergency braking system \cite{stellet2016analytical}.
To find the collision probability, 
\cite{althoff2009model} proposes to use stochastic reachable sets of states and
incorporates uncertain interactions.
Another approach that is presented in \cite{lefevre2012risk} treats a scene as risky 
if the expectated and observed intentions differ. 
The authors utilize a Dynamic Bayesian Network for this task. 
A different, control freedom-based approach is presented in \cite{constantin2014margin}. 
The proposed approach however takes uncertainties only indirectly into account; 
the one with the highest control freedom among homotopic maneuver options 
can escape from future critical situations at easiest. 

The concept of \textit{partial motion planning} in unknown, dynamic changing environments are presented in \cite{petti2005safe}. 
The work discusses the effect of perception and execution times on planning of a mobile robot. 
The earlier representives of automated vehicles utilize rule-based approaches to drive in dynamic environments. 
Calculations for complex maneuvers that are employed in the DARPA Urban Challenge are presented in \cite{werling2008robust}. 
However, the effects of visibilty and uncertainties for safe traversing are not analyzed.
A motion planner that uses probabilistic prediction and 
considers the uncertainties of localization and even control is presented in \cite{xu2014motion}. 
The planner presented in \cite{de2014collision} deals with uncertain predictions at intersections and 
considers emergency braking before intersection start. 
The work however does not consider sensor range and vehicles approaching behind the perception field. 
For planning a fail safe motion, \cite{magdici2016fail} evaluates occupancy set of the vehicles in environment 
and shows the existence of an emergency maneuver, \textit{e.g.\ }lane change. 
This work also does not take perceptive field of the vehicle into account.
The effect of visible field on planning is adressed by \cite{brechtel2014probabilistic}. 
The work uses a POMDP and is applicable to a variety of scenes. 
On the other hand, for every stiuation a new representation must be learned and 
the planner cannot perform online.  
The work presented in \cite{hubmann2017decision} also uses a POMDP for behavior planning. 
It is online, considers possible routes of others, 
but no uncertainty from visibility is taken into account and is defined only for intersections. 
An application of reinforcement learning on a similar scenario is presented in \cite{gritschneder2016adaptive}.
In that work, given observations, a linear approximator based MDP identifies discrete behavior actions 
(\textit{approach}, \textit{stop}, \textit{go} and \textit{uncertain}),  
which are subsequently sent to the motion planner. 
Limited visibility is however not investigated. 
A later work of the same research institution inspects this aspect in \cite{hoermann2017entering}. 
The work predicts occupancy probabilities with a grid map and identifies free-to-drive sections of a path. 
The work, however, does not consider the hypothesis that vehicles to which the ego vehicle must yield to, 
might appear at the limited perception horizon.

\section{Environment Model}
\label{sec:environment_model}

Behaviour and motion planning modules of an automated vehicle require an environment model as input,
that contains current spatio-temporal and relational information on the objects around the vehicle 
together with their predicted future evolution. 
 
\subsection{Localization}

Automated vehicles utilize distinct localization systems that are typically fused by a Kalman filter \cite{tas2016making}.
In our analysis, we position our ego vehicle on street relative coordinates and 
model longitudinal position and speed with a univariate Gaussian distribution 
$\mathcal{N}(\mu_x, \sigma_x^2$).
We assume position and speed measurements are uncorrelated.

\subsection{Perception}
\label{sec:environment_model:perception}

From distinct sensors percepted and fused objects are usually represented by a Gaussian distribution.
For perception, we make two assumptions:
\begin{itemize}
\item position and speed measurements are uncorrelated,
\item false negative object detection rate is 0 for a predefined range.
\end{itemize}

We further restrict our analysis to all traffic participants being vehicles. 
The approach can however be generalized for cyclists and pedestrians which in case can be modeled as dynamic boundaries.

\subsection{Scene Understanding}

Whereas spatio-temporal information is provided by sensor fusion, 
the relational information is typically extracted by scene understanding. 
For our analyis we assume the presence of an up-to-date offline map. 
The presented approach can also be extended to mapless driving 
by performing lane estimation and tracking traffic signs while considering their uncertainties.

Understanding the current scene is essential for planning behaviors and motion. 
At an intersection, whether the ego vehicle has the right of way or not must be deduced from the scene. 
In our work, we rely on the \texttt{libLanelet} digital map format \cite{bender2014lanelets} and 
extend it by tagging stop signs, speed limits, priority roads and buildings that limit visibility. 

\begin{figure}
\vspace{3mm}
\includegraphics[width=0.90\columnwidth]{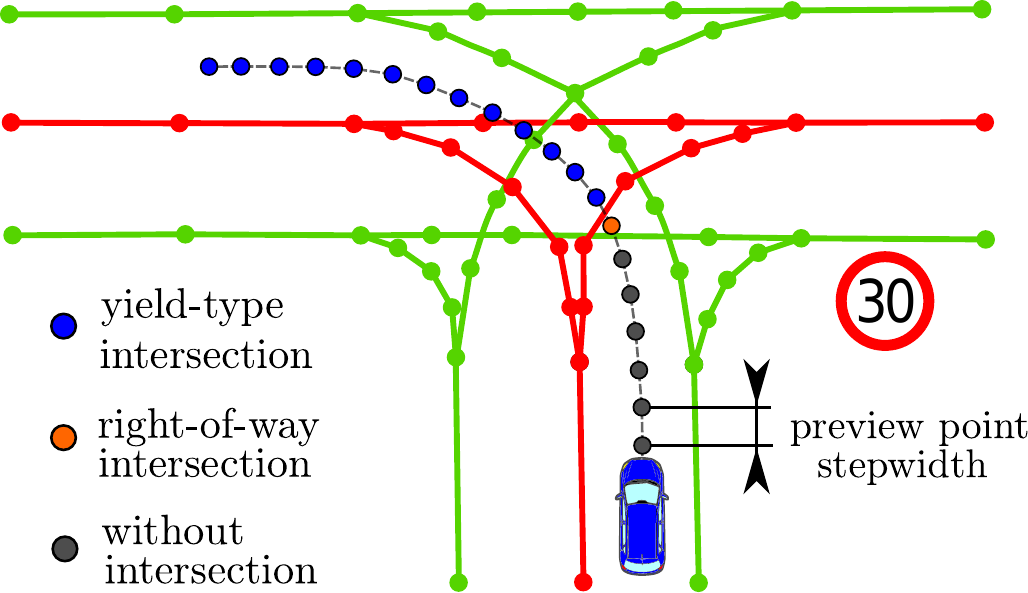}
\caption{An intersection scene, where the vehicle has to cross another road in a 30 km/h-zone.
Right-of-way at each route is determined by sampling preview points along the centerline of ego driving corridor.}
\label{fig:figure_02}
\end{figure}

\begin{figure}
\includegraphics[width=\columnwidth]{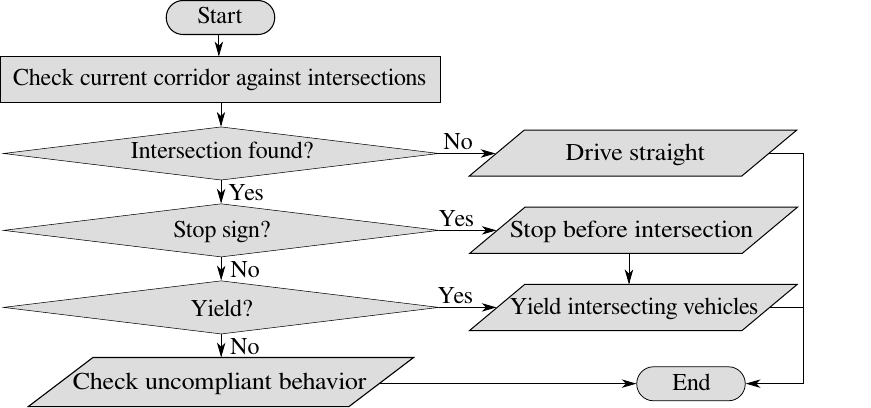}
\caption{The driving corridor is analyzed for intersections and if any is found, the action which the vehicle has to execute is determined. }
\label{fig:figure_03}
\end{figure}

For determining right-of-way, which is not a standard functionality of the \texttt{libLanelet}, 
we sample equidistant \textit{preview points} along centerline of the ego route. 
We check individual preview points whether they are 
in the vicinity of stop signs, traffic lights, 
lay to the right or to the left of an intersecting lane, 
and if they are also on other routes, \textit{cf.\ }Fig.\ \ref{fig:figure_02}.
If a point is found to be also on another route,
the driving corridor has at least one intersection. 
Using the predefined tags we determine the type of the action 
the ego vehicle must execute at that intersection, \textit{cf.\ }Fig.\ \ref{fig:figure_03}. 
A similar, state chart based approach for behavior planning is presented in \cite{gindele2008design}. 

\subsection{Situation Prediction}
\label{sec:idm}

Vehicles try to reach their destination while balancing their driving speed with comfortable and safe ride.  
The goal of prediction is to estimate the future trajectory of the traffic participants in a given scene, 
while considering their probable routes and interactions with their environment.  

In this work, we inspect traffic scenes where the route of other participants are known. 
For performing probabilistic route prediction at an intersection with multiple possible routes, 
readers are referred to e.g.\ \cite{kuhnt2016understanding} 
-- in which route probabilities are inferred from learned interactions, 
or for a broad overview to \cite{lefevre2014survey}.
The resulting multi-modal route predictions can be incorporated in planning by divide-and-conquer strategy 
likewise in \cite{constantin2014margin, bender2015combinatorial},
and afterwards picking the worst-case. 

A method to find the most likely future motion of the vehicle is to employ a motion planner that 
considers different and in case conflicting objectives and interactions with environment. 
However, instantiating a motion planner to predict the motion of all vehicles around can be computationally expensive.
We therefore utilize the well-known Intelligent Driver Model (IDM), 
which mimics human driving behavior for follow and free driving scenarios. 
Another alternative would be to use the Foresighted Driver Model \cite{eggert2015foresighted}. 
 
The acceleration profile of a vehicle on a path can be calculated as:
\begin{equation}
\label{eq:idm}
\aidm = \aacc \left( 1- {\left( \frac{v}{\vdes} \right) }^ 4 - {\left( \frac{s_\mathrm{des}}{s} + \frac{v \, \vrel}{2 \, s \, \sqrt{\aacc \,\, \acft}} \right)}^2 \right) \, ,
\end{equation}
with 
\begin{equation}
s_\mathrm{des} = \smin + v \, H_\mathrm{des} 
\end{equation}
where 
$s$ defines the non-Euclidean distance between vehicles, 
$\smin$ the stand-still distance, 
$H_\mathrm{des}$ a safe time headway,
$v$ current speed of the vehicle, 
$\vdes$ set speed of the vehicle,
$\vrel$ relative speed of the vehicle to the lead vehicle,
$\acft$ comfortable deceleration, and 
$\aacc$ maximum acceleration.
These parameters vary for different driver profiles and they can be estimated as proposed in \cite{hoermann2017probabilistic}.
However, we use fixed values as chosen in \cite{liebner2012driver}.
Note that, the IDM-based approach presented in \cite{liebner2012driver} can be employed to predict routes 
and our simplifying assumption of predefined routes can be relaxed. 

In this work as we focus on robust planning, 
we do not investigate probabilistic features of a prediction. 
For details on Gaussian uncertainty propagation of a predicted motion,
readers are referred to \cite{xu2014motion}.

\section{Robust Planning}
\label{sec:robust_planning}

The motion planner of an automated vehicle can plan a safe motion if it 
considers the uncertainties of the perception and
the worst-case evolution of the current situation,  
which involves possible uncompliant behaviors of other traffic participants. 
This requires making assumptions on how the worst-case can evolve
and defining operation modi. 

\subsection*{Notes on Motion Planner}
\label{sec:planner_notes}
Before we dive into details of our approach,   
we recapitulate our optimization-based motion planner, 
the details of which can be found in \cite{ziegler2014trajectory, tas2016making}. 
We approximate a motion $\vx(t)$ for the planning horizon $T$ by $N$
timely equidistant support points $\vx_i$, with a sampling interval of $h$. 
We utilize finite differences to find the resulting velocity, acceleration and jerk terms. 
To maintain $C_2$-continuous motion at point $\vx_{i}, i \in \{3,\ldots, N-1\}$, 
we require the support points $\vx_{i} \,,  i \in \{i_0-3,\ldots, i_0-1 \}$.
For temporal planning consistency, we pin the first
$3 + N_\mathrm{pin}$ trajectory support points. 
The $3$ points are already driven, 
whereas the $N_\mathrm{pin}$ points correspond to current time and future, 
and are being driven until the next motion is computed. 
We denote the time from which on the points are released as $t_\mathrm{pin}$
\begin{equation}
t_\mathrm{pin} = t_0 + N_\mathrm{pin}\,h \,.
\end{equation}
Note that $N_\mathrm{pin}\,h$ corresponds to \textit{dead time} $t_d$ from planning perspective.

In this paper we do not consider evasive steering maneuvers and 
limit the actions of the vehicle to acceleration and braking along a predefined path,
which we select as the mid of the driving lane.  
In this case, the planner optimizes a motion that minimizes
$J^\mathrm{d} (\mathbf{X}) \, : \mathbb {R}^{N}\, \rightarrow \mathbb {R}$, 
given the parameter vector $\mathbf{X} = (x_0, x_1, \dots , x_{N-1})^\mathsf T$.

\subsection*{Discussions on Safe Motion Planning}

Several different approaches for planning safe motion are proposed, 
as presented in Section~\ref{sec:related_work}.
To prove safety, some works propose to calculate collision probability by integrating a criticality metric-based distribution and compare it with threshold. 
This, although being reasonable, does not integrate safety in planning. 
Other approaches propose to penalize
$x_i = \{ x_i \, \mid (x_i \in \mathbb{R}) \bigwedge (x_i \in \mathbb{O}_i ) \} $ 
where $\mathbb{O}_i \sim \mathcal{N} (\mu_{x_i}^k,\, {\sigma_{x_i}^k}^2)$ is the Gaussian distributed space occupied by obstacle $k$ at time step $i$.
However, uncertainties accumulate along the horizon and even a time-gradual penalization of 
$x_i \in \mathbb{O}_i$ in uncertain regions does not guarantee safety.
This furthermore is susceptible to lead over-conservative trajectories. 

\begin{figure}[h!]
\vspace{3mm}
\includegraphics[width=\columnwidth]{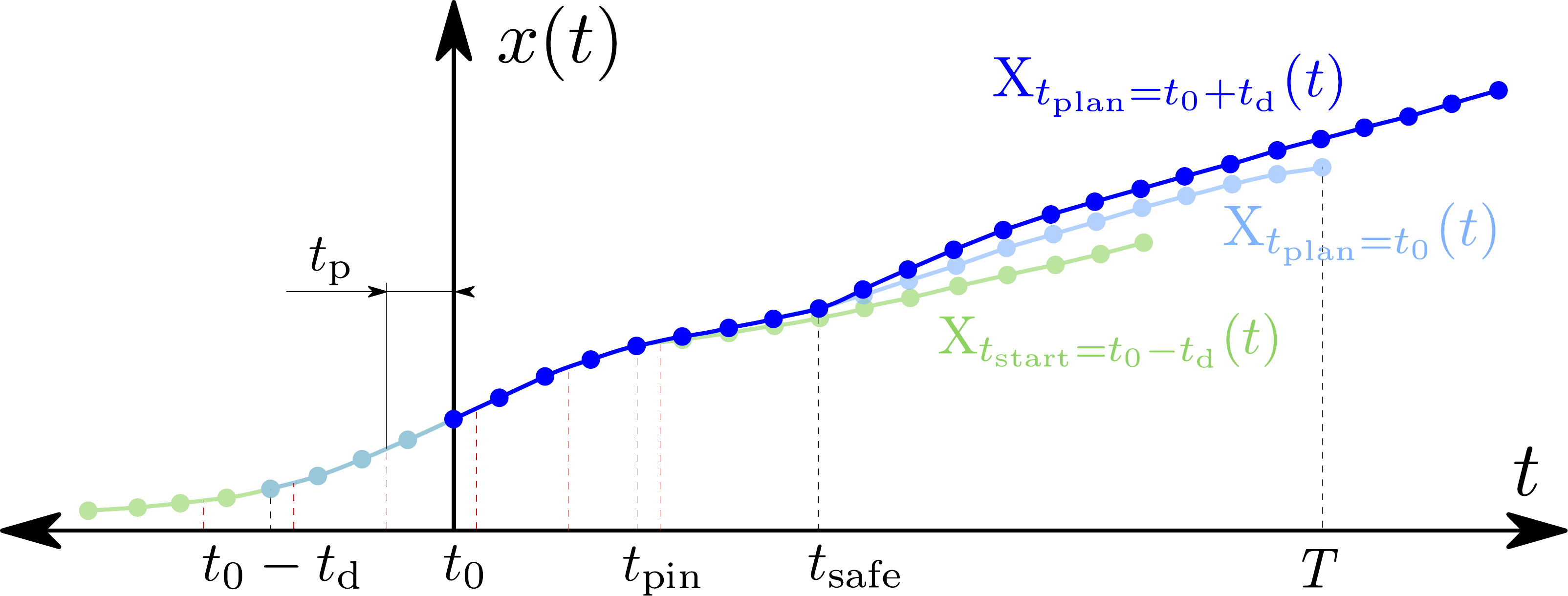}
\caption{A figure highlighting the effect of replanning and delays on motion safety. 
Three subsequent planned motions are visualized. 
The time of every new planning is visualized with black,
whereas the times at which new environment information arrive are visualized with 
red dashed lines.
The environment is updated every $2h$ time and the planner returns a new solution every $4h$. 
As a result, the planner uses $t_\mathrm{p}$ old environment information while planning. 
The motion 
$\mathrm{X}_{t_\mathrm{plan} = t_0 + t_\mathrm{d}} (t)$ can alter the motion 
$\mathrm{X}_{t_\mathrm{plan} = t_0} (t)$ only after 
$t_\mathrm{safe}$
}
\label{fig:figure_04}
\end{figure}

A planned motion must assure the existence of a safe maneuver option until 
\begin{equation}
t_\mathrm{safe} = t_0 + 2 t_\mathrm{d},
\end{equation}
given the worst-case evolution of the current scene and measurement uncertainties, 
\emph{i.e.\ }the planner must have a fallback motion option 
in the time interval $[t_0, t_\mathrm{safe}]$, \textit{cf.\ }Fig.~\ref{fig:figure_04}. 
Such a motion can be planned by defining the worst-case evolution of the scene during $2 t_d$ and 
accordingly applying additional constraints to the support points 
$x_i,  \,\, i \in \{0, \, \dots , \, 2 N_\mathrm{pin} - 1 \}$. 
Note that, planning frequency and environment model update frequency are typically not the same, 
and there is a time-shift between both.
This time is represented with $t_\mathrm{p}$ on Fig.~\ref{fig:figure_04}. 
Furthermore, environment perception process are also subject to some delay. 
Therefore, while evaluating the worst-case evolution of the scene, 
the time shift $t_\mathrm{p}$ and the perception process delay must be considered.
However, for a typical planner that is working at $50\, \mathrm{Hz}$
the error resulting from $t_\mathrm{p}$ is negligable and hence, for the sake of brevity, 
we will assume that the environment information is up-to-date. 
Moreover, if the utilized motion planner merely sends reference to low level controllers as in \cite{ziegler2014making},
and the actuation system has unnegligable delay, the delay must be added to $2 t_\mathrm{d}$.

In the following subsection we will give the definitions for worst-case evolutions and 
applied constraints for distinct scenarios.

\subsection{Straight Driving}
\label{sec:straightdriving}

As introduced in Section~\ref{sec:environment_model},
we localize the vehicle's position along a given path with the random variable $X \sim \mathcal{N} (\mu_x, {\sigma_x}^2)$
and measure its speed also with random variable $V \sim \mathcal{N} (\mu_v, {\sigma_v}^2)$.
We assume constant uncertainty in trajectory tracking,
\textit{i.e.\ }
${\sigma_{v_{i}}} = {\sigma_{v_{0}}} = \mathrm{const}$
and
${\sigma_{x_{i}}} = {\sigma_{x_{0}}} = \mathrm{const}$, $\forall i \in \{1, \ldots, N-1\}$.
We define the visibility range of the vehicle on its route as $s_\mathrm{vis}$. 
For driving along a curvy path that lacks any intersection, 
we assume that no object will interfere from the boundaries of the driving corridor and hinder the motion,
\textit{cf.\ }Fig.\ref{fig:figure_03}.
In this case, we distinguish between two modi.

\subsubsection{Free Drive}
\label{sec:freedrive}

\begin{figure}[h!]
\vspace{2mm}
\includegraphics[width=\columnwidth]{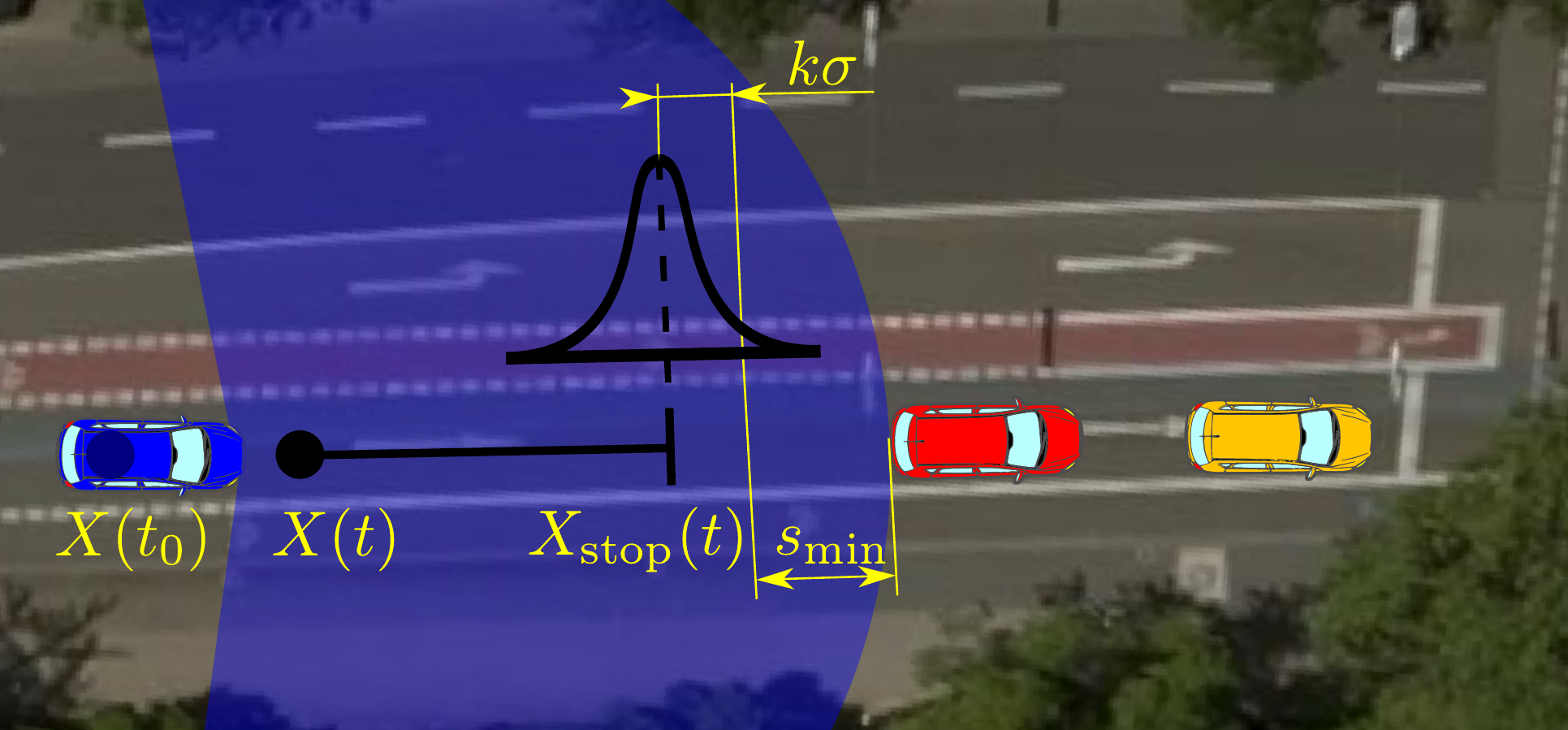}
\caption{Free driving with limited visibility.}
\label{fig:figure_05}
\end{figure}

In this mode, the vehicle is driving straight and there are not any objects in the receptive field. 
The worst-case evolution of the scene could be that, 
just behind the visible range there might be standing vehicles, 
\emph{e.g.\ }due to a traffic jam \textit{cf.\ }Fig.~\ref{fig:figure_05}. 
Note that, the worst-case can also be represented by a \textit{virtual} object,
that is stopping at the end of the receptive field. 
The braking distance $\sfull$ can be calculated as
\begin{equation}
\label{eq:braking_dist}
\sfull = \frac{v^2}{2\adec}
\end{equation}
Because $V$ is assumed to be Gaussian distributed and $\adec$ is precisely known,
the random variable ${\Sfull}$ is approximated by $\mathcal{N} (\mu_{\sfull}, {\sigma}_{\sfull}^2)$ as well.
The mean $\mu_{\sfull}$ can be calculated by using (\ref{eq:braking_dist}),
whereas the variance ${\sigma_{\sfull}}^2$ by applying 
linear approximation\footnote{For details please refer to Gaussian law of error propagation.} 
on (\ref{eq:braking_dist}):
\begin{equation}
\label{eq:variance_braking_distance}
{\sigma_{\sfull}}^2 = 4 {\sigma_v}^2.
\end{equation}
In this way ${\Sfull} (t)$ can be found for any $t \in [t_0, T]$.
Note that, a Gaussian approximation of $\adec$ would yield a Cauchy distributed ${\Sfull}$,
which in case can be approximated by a Gaussian distribution as well. 

Because $X$ and $V$ are stochastically indepent, as indicated in Section~\ref{sec:environment_model:perception}, 
the stop position 
$X_\mathrm{stop} \sim \mathcal{N} ({\mu_{x_{\mathrm{stop}}}}, {\sigma_{x_{\mathrm{stop}}}}^2)$ 
can be found by simply adding the mean and variance values of position, 
\begin{subequations}
\label{eq:braking_dist:prob}
\begin{align}
{\mu_{x_{\mathrm{stop}} i }} &= x_i + \frac{v_i^2}{2 \adec} \\
{\sigma_{x_{\mathrm{stop}} i}} &= \sqrt{ {\sigma_x}_i^2 + 4{\sigma_v}_i^2 } .
\end{align}
\end{subequations}

For a planned motion to be safe in the sense that it has a collision probability not more than $C$,
\begin{equation}
\label{eq:braking_safety}
Pr  \left\{ X(t) + \Sfull(t) \leq  X(t_0) + \svis - \smin \right\} \leq \left( 1 - C \right) , 
\end{equation}
must hold in the time interval $t \in [t_0, \tsafe]$.
For $k \in \mathbb{R}^+$, by applying (\ref{eq:braking_dist:prob}),
(\ref{eq:braking_safety}) can be reformulated as
\begin{equation}
{\mu_{x_{\mathrm{stop}} i }} + k {\sigma_{x_{\mathrm{stop}} i}} \leq \svis - \smin \,, i \, \in \{ 0, \, \ldots, 2 \npin -1 \} \,.
\end{equation}

\subsubsection{Follow Drive}
\label{sec:followdrive}

In this modus, the ego vehicle is following the most important object in the receptive field. 
We denote the variables of ego vehicle with the superscript $e$ 
and those of the percepted vehicle with $o$. 

For a planned motion to have a maximum collision probability of $C$ in a follow drive,   
\begin{equation}
\label{eq:follow_safety}
Pr \big\{ X^e (t) + S_\mathrm{full}^e (t) \leq X^o (t_0) - \smin + S_\mathrm{full}^o (t_0) \big\} \leq \left( 1 - C \right) ,
\end{equation}
must hold for the time interval $t \in [t_0, t_\mathrm{safe}]$. 
Notice that, this corresponds to the worst case at which the leader vehicle 
applies full braking immediately after the environment information is sent to the planner. 

By applying error propagation law, 
the mean value and the standard deviation can be calculated as
\begin{subequations}
\label{eq:follow_drive:prob}
\begin{align}
\mu_{x_\mathrm{follow} i} & = {x^e_i} +  \frac{{v^e_i}^2}{2\adec} \\
\sigma_{x_\mathrm{follow} i} & = \sqrt{ {\sigma^e_{x_i}}^2 + {\sigma^o_{x_0}}^2 + 4{\sigma^e_{v_i}}^2 + 4{\sigma^o_{v_0}}^2 } 
\end{align}
\end{subequations}
The parameters $\sigma^o_{x_0}$ and $\sigma^o_{v_0}$ represent the standard deviation of 
position and speed measurements of the percepted vehicle at $t_0$.
By applying (\ref{eq:follow_drive:prob}) on (\ref{eq:follow_safety}),  
the constraint to be applied on the motion planner is obtained
\begin{align}
\mu_{x_\mathrm{follow} i} + k {\sigma_{x_\mathrm{follow} i}} \leq {x^o_0}  & - \smin + \frac{{v^o_0}^2}{2\adec} \,, \nonumber \\
& i \, \in \{ 0, \, \ldots, 2 \npin -1 \} \,.
\end{align}

Note that, in contast to \cite{ziegler2014trajectory},
we add a further summand
\begin{equation}
j_{s} = w_{\mathrm{s}} \abs{s(t)_\mathrm{des} - s(t)}^2
\end{equation}
to the objective functional for maintaining an ideal follow behavior, 
as presented in \cite{tas2016making}.

\subsection{Intersection Crossing}

If the automated vehicle is approaching to an intersection, 
when the horizon of the visible area reaches to the merge point (\mergepoint) 
while there are no other vehicles driving in front,
the planner switches to the intersection crossing mode.
In order to systematically deal with the scenario,
we make further assumptions: 
the vehicles in the environment do not exceed speed limits 
and the reachable maximum deceleration $\adec$ for the vehicles in the intersecting route is known. 

\begin{figure}[h!]
\includegraphics[width=\columnwidth]{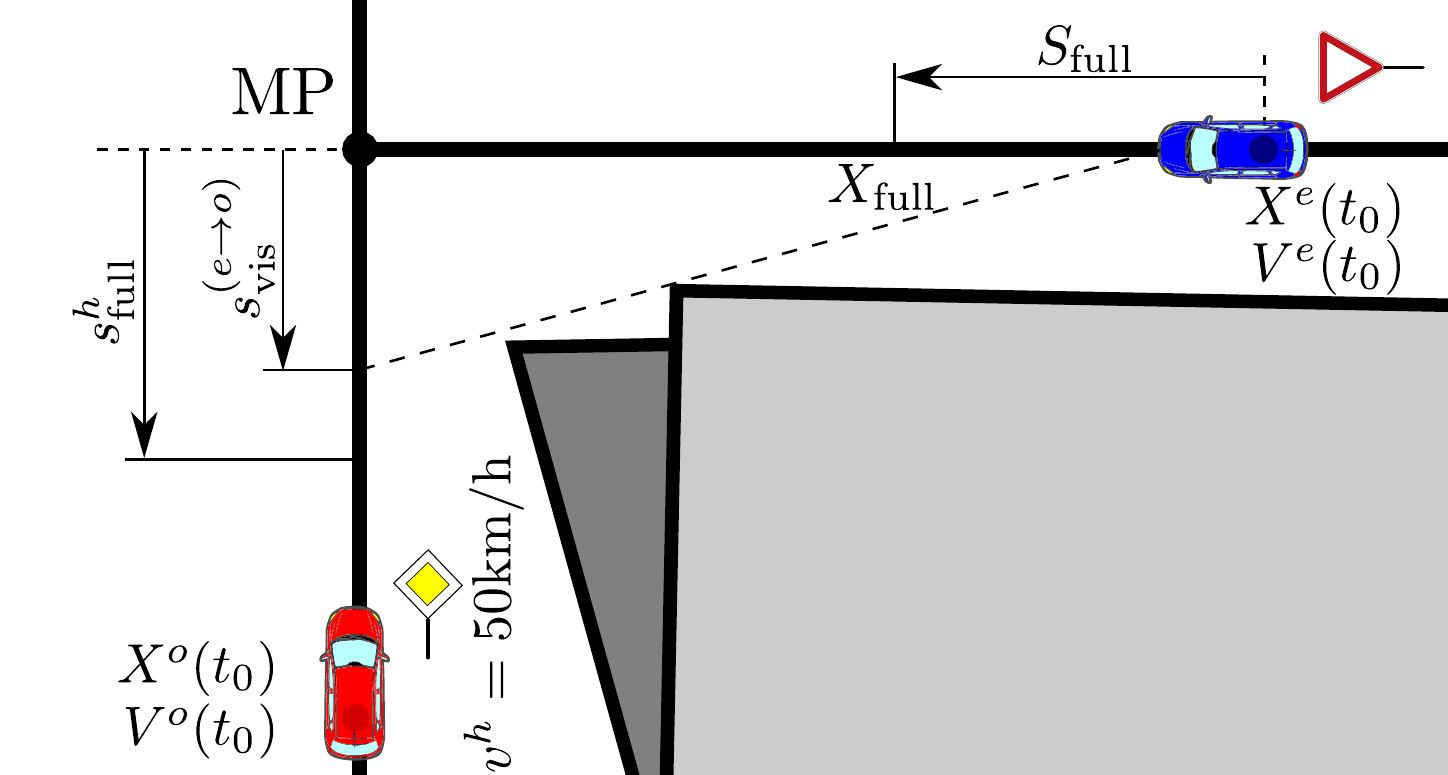}
\caption{Two paths cross at the merge point $\mathrm{MP}$. 
The paths are not neccessarily straight, and hence the distances are non-Euclidean.
The visibility of the ego vehicle is currently hindered by the grey polygon. 
However, at a later time, the visibility might be hindered by another currently unseen polygon,
\textit{e.g.\ }by the hatched one.}   
\label{fig:figure_06}
\end{figure}

Visible areas are calculated by assuming the polygons limiting the visibility have a convex shape. 
Given road geometry and the convex polygons in the environment, 
the farthest visible point for the ego vehicle on intersecting route 
can be calculated using computational geometry. 
We denote the distance between this point and \mergepoint $ $ with $\svisproj$,
\textit{cf.\ }Fig.~\ref{fig:figure_06}.

Inside the invisible region there might be a \textit{hypothetical} vehicle approaching to the intersection. 
By assuming that the vehicle will obey the speed limit, 
we can assign the speed limit for that route as the speed $v^h$ of the hypothetical vehicle. 
The braking distance for the hypothetical vehicle for full $\adec$ can be denoted with $s_\mathrm{full}^h$.
It should be underlined that
the hypothetical vehicle is at the upper bound of speed and its position is also well-defined.
The hypothetical vehicle furthermore corresponds to the MIO for that part of the driving route. 

Depending on the scenario, we distinguish between two maneuvers,
which we analyze in the following.

\subsubsection{Give-Way Maneuvers}

For give-way maneuvers, we check whether the vehicle detects a vehicle within $\svisproj$. 
If there is no vehicle, 
then we assume that the most important object (MIO) is a hypothetic vehicle 
that is travelling with $v^h$ at the line of sight. 
If there is a vehicle detected, 
we assume as the worst case scenario that the vehicle is followed by a hypothetical vehicle 
at the line of sight. 

In order to imitate interactions and to determine whether 
the ego vehicle can enter intersection without interfering the MIO,  
we refer to the IDM, presented in Section~\ref{sec:idm}. 
A conservative prediction from the perspective of the MIO 
will assume the speed of the ego vehicle to be constant. 
With the insights from the MOBIL \cite{kesting2007general},
if the neccessary acceleration $\aidm^o$ to react to ego vehicle satisfies
\begin{equation}
\label{eq:drivein}
\lvert \aidm^o (t) \rvert  \leq (1 - \gamma^e) \lvert  \acft^o \rvert  \, , \forall t \in  [t_0, T] 
\end{equation}
where $\gamma^e \, \in [0, 1)$ is the politeness factor of the ego vehicle. 

\begin{figure}[h!]
\vspace{3mm}
\includegraphics[width=\columnwidth]{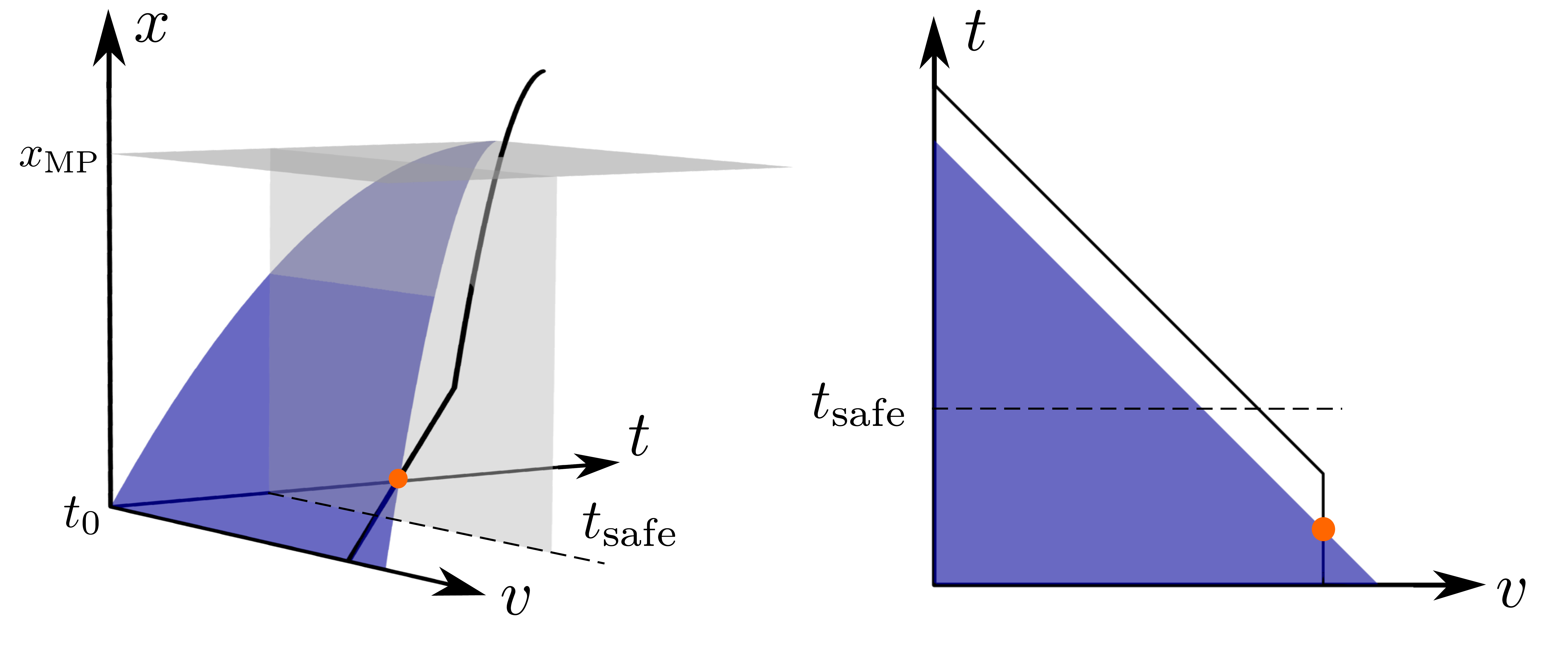}
\caption{
Safe motion in path--time--speed $x \times t \times v$ space for the intersection scenario. 
Any motion that is below the blue surface can come to full stop before the intersection. 
We call the surface of this volume as \textit{surface-of-no-return}. 
Any point on this surface, such as the orange point, is the \textit{point-of-no-return} 
which is mentioned but not precisely defined in \cite{gindele2008design}, \cite{de2014collision}.
For a planned motion to be considered as safe, 
it must be below the blue surface until $t_\mathrm{safe}$. 
If it leaves the surface afterwards, it can be recovered in the next planning iteration. 
The in black depicted motion is in this sense \textit{unsafe}. 
However, it may still be collision free, as this depends on the motion of the other vehicle as well.}
\label{fig:figure_07}
\end{figure}

The presented naive interaction check is done for both of the vehicles. 
A systematic analysis for combinatorial alternatives is presented in \cite{bender2015combinatorial, altche2018partitioning}. 
If the ego vehicle can drive into intersection without interfering MIO, 
we do not apply further constraints. 
The vehicle must only deal with \textit{Free Drive} case, 
presented in Section~\ref{sec:freedrive}. 
If the vehicle should give way to MIO and does not interfere the follower of the MIO,
which may be the hypothetic vehicle,
it must deal with the \textit{Follow Drive} case, 
presented in Section~\ref{sec:followdrive}. 
Otherwise, if the vehicle must yield to MIO and the visibility behind the MIO is not sufficient to merge in,
\begin{equation}
\mu_{{x_\mathrm{stop}}_i} + k \sigma_{{x_\mathrm{stop}}_i}  \leq \xMP - \smin \, , i \in \{0, \, \ldots, \, N -1\} .
\end{equation} 
must hold for the entire planning horizon. 
Note that $i \in \{0, \, \ldots, \, 2N_\mathrm{pin} - 1 \}$ would also be sufficient, 
\textit{cf.\ }Fig.~\ref{fig:figure_07} for a discussion.

\subsubsection{Right-of-Way Maneuvers}

An automated vehicle 
should be able to compensate uncompliant behavior of other traffic participants,
when approaching to an intersection where it has the right-of-way.
While being proactive, it should not move too defensive. 
Its intended motion must be 
transparent\footnote{We define \textit{transparency} of a planned motion as \emph{how well} the intended maneuver can be perceived by other traffic participants.} 
enough to reflect that it intends to preserve its right-of-way. 
The both goals are contradictory and hence, we aim to apply brakes at $\text{Time-to-Brake} = 0$.

Our planning approach can deal with non-compliance for the cases where 
the MIO does not adapt its speed to visibility and
to where the MIO does not start braking comfortably to yield to the ego vehicle. 
For the visibility case we check whether 
\begin{equation}
\label{eq:incompliantvisibility}
\svisproj > \sfull^h + 2 t_\mathrm{d} v^h. 
\end{equation}
If this does hold and there is a vehicle in the perception field, 
we analyze if the vehicle is uncompliant in the same way as done for merging in \textit{Give-Way} maneuvers. 
We check whether the neccessary acceleration $\aidm^o$ to give way to ego vehicle 
exceeds comfortable braking deceleration $\acft$. 
If 
\begin{equation}
\label{eq:incompliantdeceleration}
\lvert \aidm(t) \rvert > \lvert \acft \rvert \, , \forall t \in  [t_0, T]
\end{equation}
the vehicle is assumed to violate the rules. 
In case any of the inequalities (\ref{eq:incompliantvisibility}) and (\ref{eq:incompliantdeceleration}) do not hold, 
we ensure that the planned trajectory of the ego vehicle holds
\begin{equation}
\mu_{{x_\mathrm{stop}}_i} + k \sigma_{{x_\mathrm{stop}}_i} \leq \xMP - \smin \,, \,\, i \in \{0, \, \ldots, \, 2 \npin -1\} .
\end{equation} 
  
In the presented scenarios, 
by applying the presented constraints on the optimization-based motion planner, 
comfortable and safe motion profiles can be obtained.  

\section{Experiments}
\label{sec:experiments}

The proposed planner is tested in a closed-loop simulation environment developed by the authors. 
The optimization problem, the basics of which are recapitulated in Section~\ref{sec:planner_notes}, 
with the application of defined constraints are solved with the optimization library Ceres \cite{ceres-solver}. 
For calculating gradients and Hessians, the automatic differentiation of Ceres is utilized.  
Thereby, the burden of calculating analytical derivatives by using symbolic representations 
is resolved with the utilization of the chain rule. 
As Ceres does not support \textit{hard} constraints, 
we approximate these by applying barrier methods as further cost terms.

\begin{figure}[h!]
\vspace{2mm}
\includegraphics[width=\columnwidth]{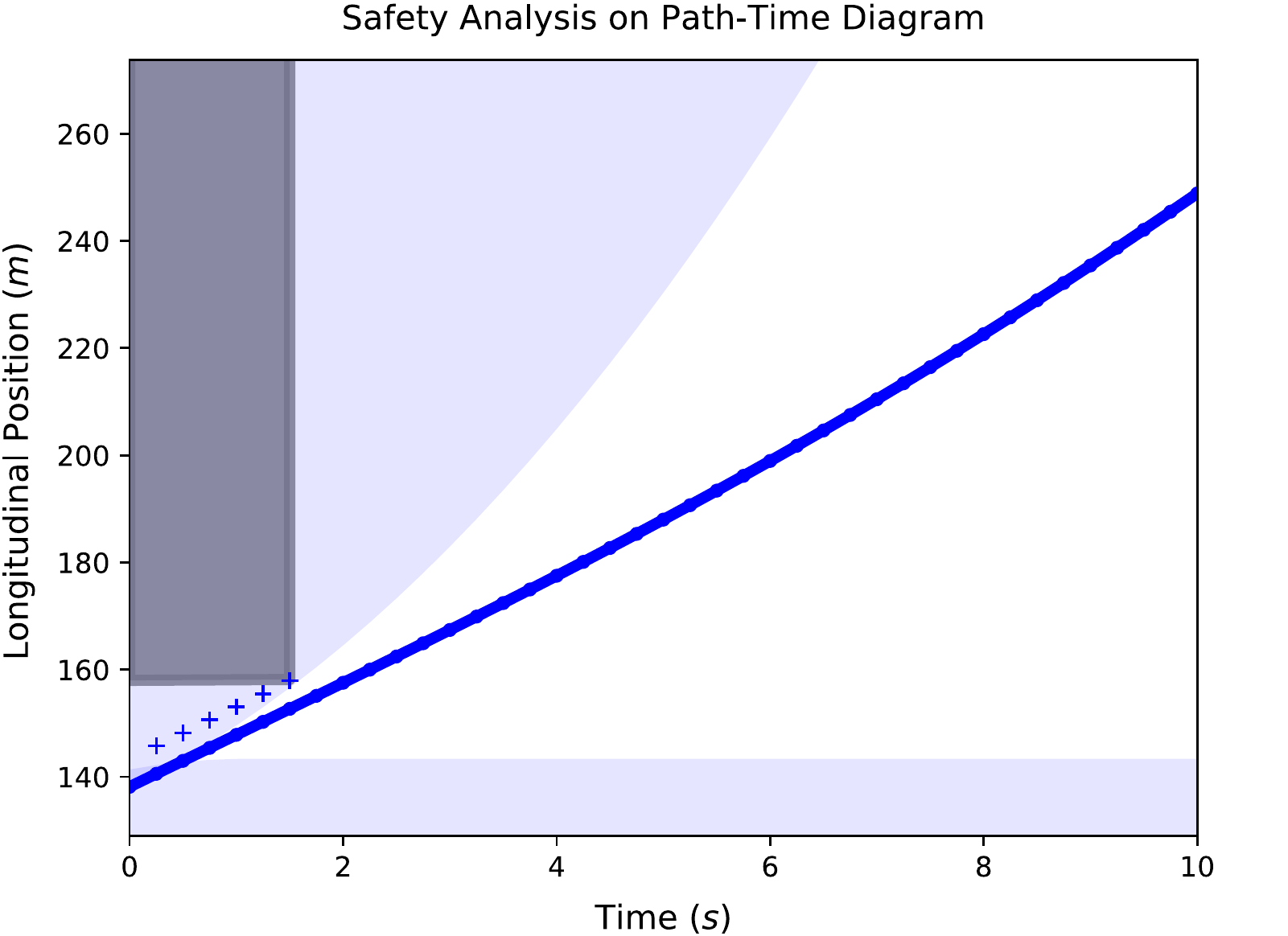}
\caption{Safety analysis on path-time diagram.
The blue regions represent the states that are not reachable by the vehicle. 
The gray region represent the states that are occupied by the hypothetic vehicle. 
The thick blue line corresponds to the optimized motion, 
whereas with "+" denoted points are the full braking stop points 
of the trajectory support points of the same time-index. 
The sampling interval $h$ is $250 \mathrm{ms}$ and $N_\mathrm{pin}$ is $3$.
}
\label{fig:figure_08}
\end{figure}

For a free driving scenario with limited visibility, 
a safety analysis on path-time diagram is given in Fig.~\ref{fig:figure_08}.
The projection of the hypothetical vehicle on path-time diagram is depicted with gray. 
All of the trajectory support points within $t \in [t_0, t_\mathrm{safe}]$ have stop distances lower than the visible range. 

\begin{figure}
    \centering
    
    \begin{subfigure}[b]{0.49\columnwidth}      
        \includegraphics[width=\columnwidth]{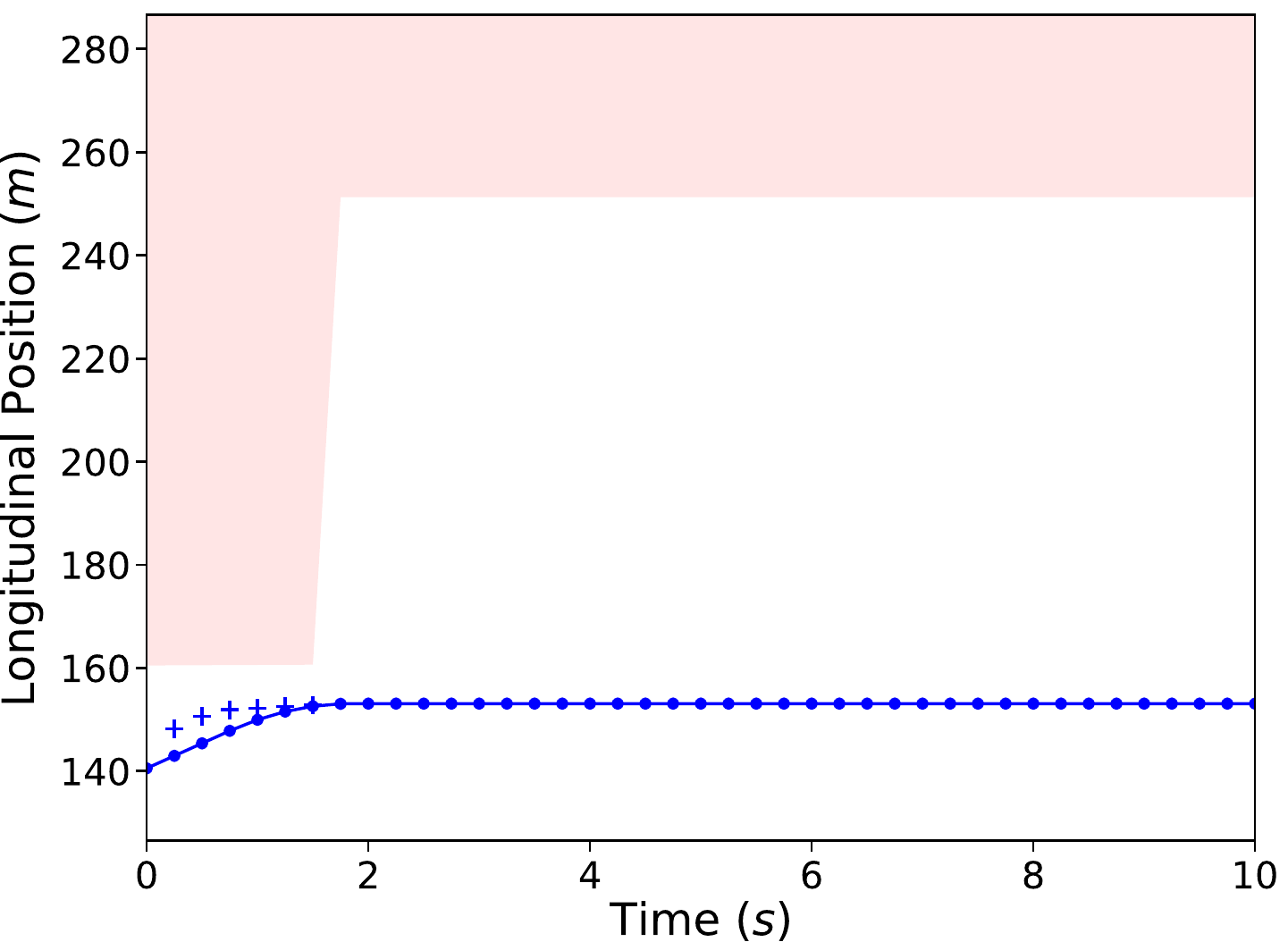}
        \caption{Initialization.}
        \label{fig:initialization}
    \end{subfigure}
    \begin{subfigure}[b]{0.49\columnwidth}      
        \includegraphics[width=\columnwidth]{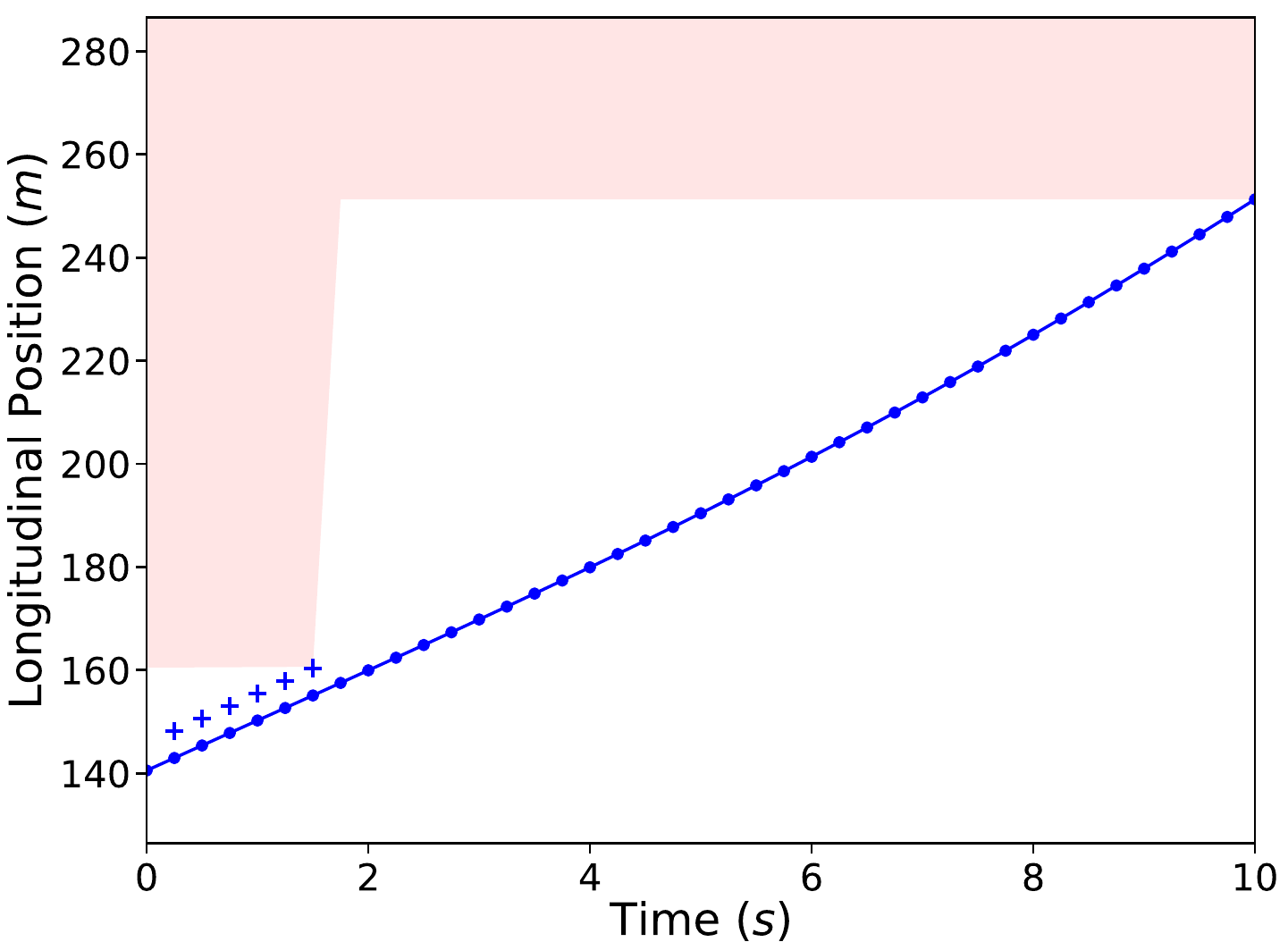}
        \caption{Solution.}
        \label{fig:optimization}
    \end{subfigure}
    
    \caption{Initialization and the result of the optimization process. 
    The red regions correspond to the hard constrained states. 
    The upper bound is calculated by multiplying speed limit with the planning horizon $T$.
    Notice that, the points after $N_\mathrm{pin}$ are altered, 
    whereas points until $2 N_\mathrm{pin}$ are bound to satisfy the hypothetical vehicle constraints.}
    \label{fig:figure_09}
\end{figure}

Because the planner is based on local optimization,
an initial guess must be provided to the solver. 
We calculate this by taking the first $N_\mathrm{pin}$ terms of the previously optimized solution 
and performing full braking for the remaining points in the planning horizon,
\textit{cf.\ }Fig.~\ref{fig:figure_09}. 
In this way, even in the case where the solver fails to find a solution, 
the planner will guide the vehicle to a \textit{safe} stop. 

\begin{figure}[h!]
\vspace{2mm}
\includegraphics[width=\columnwidth]{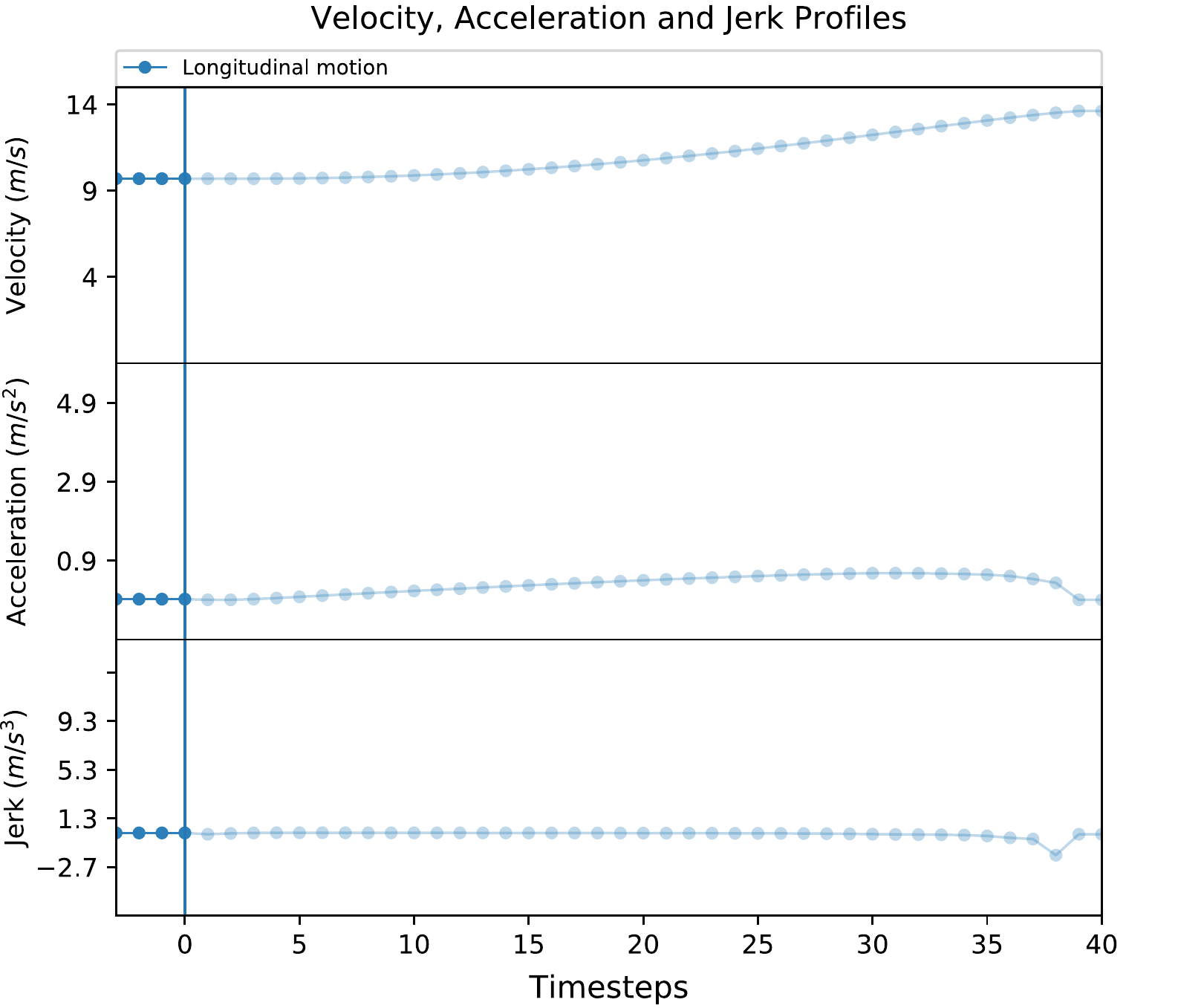}
\caption{The profile of the optimized motion. 
The quality of motion can be inferred from the smoothness of the individual profiles. 
The jump at the end of the planning horizon is due to the utilization of forward finite differences.}
\label{fig:figure_10}
\end{figure}

The resulting motion profile is shown in Fig.~\ref{fig:figure_10}.
Even though the desired travel speed is $13.89 \, \mathrm{m/s}$, 
the vehicle drives with $9.69 \, \mathrm{m/s}$ in order to satisfy the constraints arising from limited visibility. 

\begin{figure}[h!]
\vspace{2mm}
\includegraphics[width=\columnwidth]{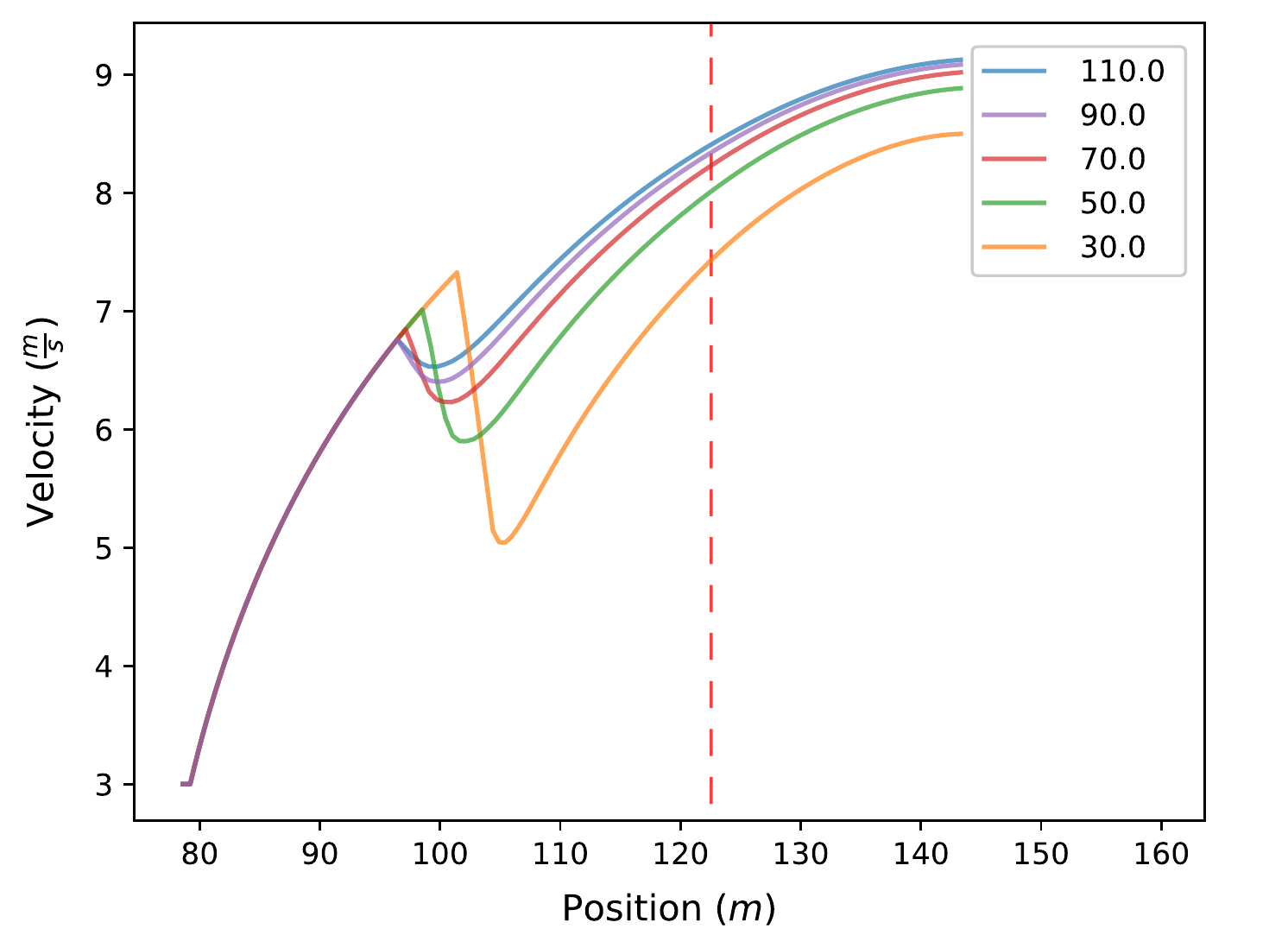}
\caption{For various sensor ranges ($30\mathrm{m} -110 \mathrm{m}$) the speed profile along the path of an intersection scenario.
The vertical dashed red line corresponds to the merge point $\mathrm{MP}$.}
\label{fig:figure_11}
\end{figure}

Simulations for intersection crossing have also been performed. 
The effect of sensor range on the speed profile for an intersection scenario, 
such as the one presented in Fig.~\ref{fig:figure_01}, 
is presented in Fig.~\ref{fig:figure_11}.  
The vehicle starts to accelerate and approaches to the intersection. 
For high sensor ranges, the preview points that are visualized in Fig.~\ref{fig:figure_02} are longer 
and hence, the intersection is detected earlier. 
This leads the vehicle to decelerate gently. 
Because the vehicle can perceive a more broad area for longer sensor ranges, 
it can pass the intersection by only slightly reducing its speed. 
However, for shorter sensor ranges, the intersection is detected later. 
Because the distance which the vehicle can detect on the intersecting route is shorter, 
the vehicle substantially reduces its speed to adapt its braking distance. 
The effect of visible field can also be observed from the final speed after the intersection: 
longer horizons lead to a more broad visible area to which the vehicle is driving to.
This, in return, relaxes the hypothesis on the presence of a stopping vehicle right after the intersection,
and eventually allows the vehicle to turn the intersection faster.
However, it should be underlined that the presented characteristics on Fig.~\ref{fig:figure_11} heavily 
depend on the structure of the environment. 
If the visibility of an intersection zone is hindered by the structures in the environment, 
the effect of sensor range can hardly be observed. 

Apart from the results presented here,  
several further scenarios are tested in simulation\footnote{
The readers are kindly requested to visit \footnotesize{\url{https://url.fzi.de/tas2018limited}} to view 
more comprehensive simulation results.}:
the effect of uncertain perception and localization, and politeness factor is inspected as well. 
As expected, increasing uncertainties lead to longer distances between vehicles, 
and reducing politeness factor leads to more aggressive driving maneuvers.

\section{Conclusions and Future Work}
\label{sec:conclusion}

In this paper we analyzed uncertainties that an automated vehicle is subject to. 
We studied different challenging traffic situations for a vehicle with a limited receptive field. 
We presented conditions for motion planner 
to consider vehicles approaching from unvisible regions to which must be yielded.  
Even for the case at which the automated vehicle has the right-of-way,
we derived an approach to detect uncompliant behaviors. 

The results imitate how human drivers approach intersections. 
For shorter sensor ranges the automated vehicle drives with reduced speed
This emulates driving in bad conditions, 
\emph{i.e.\ }sensor degradation or even partial sensor failures. 
In this sense, the results obtained from the continuous optimization based planner
imitates exploration step of a MDP. 
Because the planner is based on continuous formulations, 
there are not any inherent discretizations as well.

The proposed approach reflects the architectural guidelines of the \textsc{RobustSENSE} project:
the uncertainties are propagated up to the final layer and are treated together with collision probabilities \cite{tas2017automated}. 
The results will be demonstrated during the \textsc{RobustSENSE} project final event on May 16th, 2018 in Ulm, Germany. 
The presented concepts are currently integrated in the automated vehicle \textsc{BerthaOne} \cite{tas2016making}. 
The simulated visible range is computed from an occupancy grid map.
Our future work will deal with more complex scenarios, 
where a multitude of routes and manuever options such as lane changes are available.

\section*{Acknowledgements}

The research leading to these results has received funding 
from the European Union under the H2020 EU.2.1.1.7.\ \textsc{ECSEL} Programme, 
as part of the \textsc{RobustSENSE} project, contract number 661933.
Responsibility for the information and views set out in this publication lies entirely with the authors. The authors would like to thank all partners for
their cooperation and valuable contribution.

\bibliographystyle{IEEEtran}
\bibliography{tas2018limited}

\end{document}